\documentclass[10pt,twocolumn,letterpaper]{article}

\usepackage{cvpr}              

\usepackage[dvipsnames]{xcolor}

\usepackage{times}                     
\usepackage{amsmath}
\usepackage{amssymb}
\usepackage{multirow}
\usepackage{adjustbox}
\usepackage{graphicx}
\usepackage{gensymb}
\usepackage{subcaption}
\usepackage{cite}                      
\usepackage{booktabs}                  
\usepackage{color}
\usepackage{xcolor}

\usepackage[normalem]{ulem}

\definecolor{linkcolour}{rgb}{0.8,0,0.8}
\definecolor{citecolour}{rgb}{0,0.6,0.2}
\definecolor{urlcolour} {rgb}{0,0.2,0.8}
\definecolor{orange}{RGB}{250,140,0}
\definecolor{darkgreen}{RGB}{0,150,0}

\newcommand{\blue}[1]{\textcolor{blue}{#1}}

\newcommand\customparagraph[1]{\vspace{0.4em}\noindent\textbf{#1}}

\definecolor{cvprblue}{rgb}{0.21,0.49,0.74}
\usepackage[pagebackref,breaklinks,colorlinks,citecolor=cvprblue]{hyperref}

\def\paperID{148} 
\def\confName{3DV\xspace}
\def\confYear{2024\xspace}

\title{Improved Scene Landmark Detection for Camera Localization}

\author{Tien Do\thanks{work done while Tien Do was affiliated with Microsoft.}\\
Tesla
\and
Sudipta N. Sinha\\
Microsoft
}

\begin{document}
\maketitle
\begin{abstract}
Camera localization methods based on retrieval, local feature matching, and 3D structure-based pose estimation are accurate but require high storage, are slow, and are not privacy-preserving. A method based on scene landmark detection (SLD) was recently proposed to address these limitations. It involves training a convolutional neural network (CNN) to detect a few predetermined, salient, scene-specific 3D points or landmarks and computing camera pose from the associated 2D--3D correspondences. Although SLD outperformed existing learning-based approaches, it was notably less accurate than 3D structure-based methods. In this paper, we show that the accuracy gap was due to insufficient model capacity and noisy labels during training. To mitigate the capacity issue, we propose to split the landmarks into subgroups and train a separate network for each subgroup. 
To generate better training labels, we propose using dense reconstructions to estimate visibility of scene landmarks. Finally, we present a compact architecture to improve memory efficiency. Accuracy wise, our approach is on par with state of the art structure-based methods on the \textsc{Indoor-6} dataset but runs significantly faster and uses less storage. Code and models can be found at \url{https://github.com/microsoft/SceneLandmarkLocalization}.
\end{abstract}
\section{Introduction}

In this paper, we study the task of estimating the 6-dof camera pose with respect to a reconstructed 3D model of a scene from a single image. This is an important task in robotics and augmented reality applications. The most common approach for solving the task is structure-based~\cite{sarlin2019,sarlin2020superglue,sattler2012aachen,sattler2012improving}, where typically, the local 2D image features are matched to 3D points in a scene model. Geometric constraints derived from the 2D–3D matches are then used to compute the camera pose. These methods can be quite accurate but the need to persistently store a lot of features and 3D points raises privacy issues~\cite{pittaluga2019} and also makes them less suitable for resource-constrained settings.

Learning-based localization methods~\cite{kendall2015,brachmann2017dsac,brachmann2021dsacstar,Pietrantoni_2023_CVPR} can alleviate both the storage and privacy issues. However, despite much progress on learning-based localization, most of the methods are still not competitive with structure-based methods~\cite{sarlin2019,sarlin2020superglue}. Recently, Do \etal \cite{Do2022} proposed SLD, a localization framework that involves training CNNs for detecting pre-selected, scene landmarks (3D points) and regressing 3D bearing vectors (NBE) for the landmarks. The 2D detections and 3D bearing predictions are jointly used (SLD+NBE) to compute camera pose. Even though SLD+NBE outperforms learning-based methods~\cite{kendall2015,brachmann2021dsacstar} on the challenging \textsc{Indoor-6} dataset, it is less accurate than hloc~\cite{sarlin2019,sarlin2020superglue} by a notable margin. It is also unclear to what extent the method can handle a large number of landmarks.

\begin{figure*}
  \includegraphics[width=\linewidth]{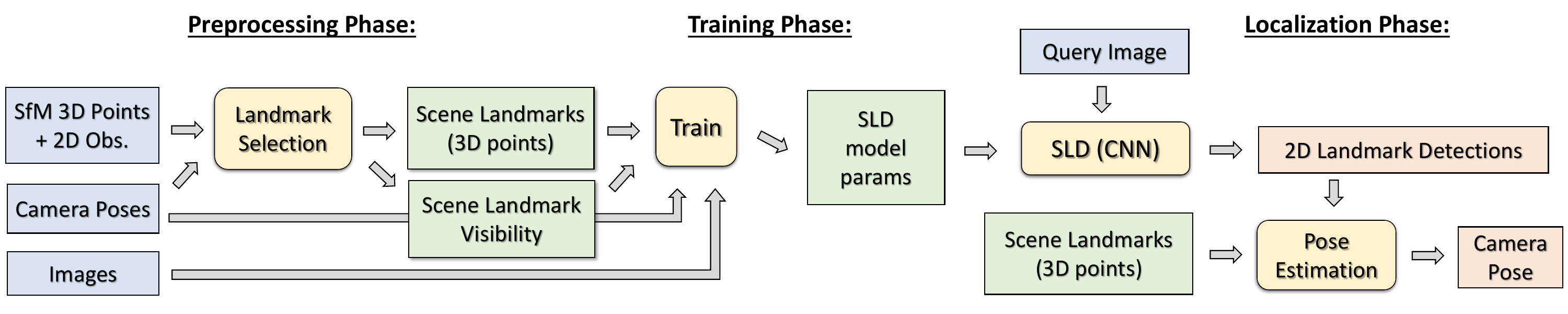}
  \caption{Key elements of the scene landmark detection-based localization approach~\cite{Do2022}. The figure shows a single model (SLD) for brevity, but Do \etal~\cite{Do2022} also proposed predicting landmark bearings using an additional model (NBE). This is discussed in the text.}
  \label{fig:overview}
  \vspace{-2mm}
\end{figure*}

In this paper, we present important insights into what typically hurts SLD's accuracy and scalability. Our first finding is that insufficient model capacity is a key cause for a drop in performance when SLD is trained for a larger set of landmarks. We also find that the automatic structure from motion (SfM) processing phase which generates labeled training patches for landmarks from training images can produce erroneous training labels. Such outliers can sometimes affect the accuracy of models trained on the data. 

To address the capacity issue, we propose to partition the set of scene landmarks into mutually exclusive subgroups, and train an ensemble of networks, where each network is trained on a different subgroup. Using an ensemble improves accuracy for scenes where a larger number of landmarks are present. To reduce the amount of erroneous labels in the training set, we propose using a dense scene reconstruction to recover more accurate visibility estimates of the scene landmarks in the training images, especially under strong lighting changes. We show that better training labels leads to more accurate landmark detections. We also propose SLD\textsuperscript{$\ast$}, a variation of the SLD architecture that improves memory efficiency, and explore using output prediction scores as a confidence measure during pose estimation.

Incorporating all the proposed ideas leads to a dramatic improvement in pose estimation accuracy, making SLD\textsuperscript{$\ast$} competitive with hloc on the \textsc{Indoor-6} dataset. At the same time, it is also \emph{more than 40$\times$ faster} than hloc during localization and \emph{20$\times$ more storage efficient}. Furthermore, SLD\textsuperscript{$\ast$} is 20--30\% more memory efficient than SLD.

\section{Related Work}

\customparagraph{Structure-based Localization.}
Classical structure-based approaches use pre-computed 3D scene point clouds to compute camera pose by combining efficient visual retrieval~\cite{asmk,sattler2012aachen,torii2019large,arandjelovic2016netvlad,pion2020benchmarking}, feature matching~\cite{sattler2012improving,lim2015,revaudr2d2,densevlad,detone2018superpoint,sarlin2020superglue}, and geometric pose estimation\cite{Ke_2017_P3P}. 
hloc ~\cite{sarlin2019} is such a method with state-of-the-art performance on \textsc{Indoor-6} that uses learning for more accurate feature matching~\cite{revaudr2d2,detone2018superpoint,sarlin2020superglue,NIPS2017_831caa1b}.
While correspondences and pose is usually estimated independently, jointly refining deep multiscale features and camera pose has been shown to improve accuracy~\cite{sarlin21pixloc}. 
Alternatively, retrieval-based methods~\cite{arandjelovic2016netvlad,asmk,densevlad} can estimate the camera pose by interpolating 
poses of retrieved database images\cite{torii2019large}. Efficient and scalable alternatives for large-scale location
classification and place recognition have also been studied~\cite{bergamo2013leveraging,gronat2013,weyand2016}.

\customparagraph{Learning-based Localization.}
Learning-based techniques do not require storing 3D scene models. 
A popular approach is to train models to regress the camera pose directly from the query image, which is called absolute pose regression (APR). 
PoseNet~\cite{kendall2015} first proposed end-to-end trainable CNN architectures, which have been extended for leveraging attention mechanisms~\cite{wang2020atloc} and to use transformer architectures~\cite{shavit2021learning}. However, APR methods rely on training sets with homogeneous camera pose distributions. When the pose distribution is highly heterogeneous, performance can suffer on such datasets~\cite{sattler2019understanding}, as was reported on \textsc{Indoor-6} in~\cite{Do2022}. Unlike APR approaches, relative 
pose regression (RPR) approaches predict the relative pose with respect to stored database images~\cite{laskar2017iccv,baintas2018eccv}. They usually generalize better but have higher storage costs.

\customparagraph{Scene Coordinate Regression.}
In contrast to APR and RPR methods, scene coordinate regression (SCR)~\cite{shotton2013scene} approaches involve training model that predict dense 3D coordinates for points in the query image and computing pose from the dense 2D--3D correspondences. DSAC\cite{brachmann2017dsac} was amongst the earliest works to 
propose an end-to-end differentiable SCR architecture. Subsequently, the framework was extended for improved efficiency during inference~\cite{li2018fullframe}, removing the need for RGBD ground truth during training~\cite{brachmann2018lessmore}, and improving the accuracy and robustness of the underlying method~\cite{brachmann2021dsacstar}. 
Other ideas have been explored, such as, the use of ensembles to improve scalability \cite{brachmann2019esac}, design of hierarchical scene representations~\cite{li2020hscnet} and scene agnostic approaches\cite{yang2019sanet}, and ideas to make the models amenable to continual updates~\cite{Wang2021} and faster training~\cite{BrachmannACE2023}.

Finally, we review methods for privacy-preserving localization, and storage efficiency. Speciale et al~\cite{speciale2019a} explored new geometric scene and query representations~\cite{speciale2019a,speciale2019b} and proposed pose estimation techniques for those representations. GoMatch~\cite{zhou2022} is a storage efficient method for geometric matching of 2D keypoints and 3D points that does not require local descriptors. SegLoc~\cite{Pietrantoni_2023_CVPR} achieves storage efficiency by leveraging semantic segmentation-based map and query representations. Approaches leveraging objects of interests in the scene~\cite{weinzaepfel2019visual} have also been studied. 
\section{Proposed Methodology}

We now present a brief review of SLD before describing our proposed ideas for improvements, in the following sections.

\subsection{Background: Scene Landmark Detection}

Do \etal~\cite{Do2022} proposed a localization approach where given the SfM reconstruction of the mapping images, a few salient, scene-specific 3D points are first selected from the SfM point cloud. Then, two CNNs (SLD, NBE) are trained using the mapping images and their associated poses. While SLD detects the  landmarks visible in images, NBE regresses 3D bearing vectors for all the landmarks in the scene. Finally, the 2D--3D landmark constraints are used to recover the camera pose. Figure~\ref{fig:overview} provides an overview.

\begin{figure*}[t]
  \centering
  \includegraphics[width=0.86\linewidth]{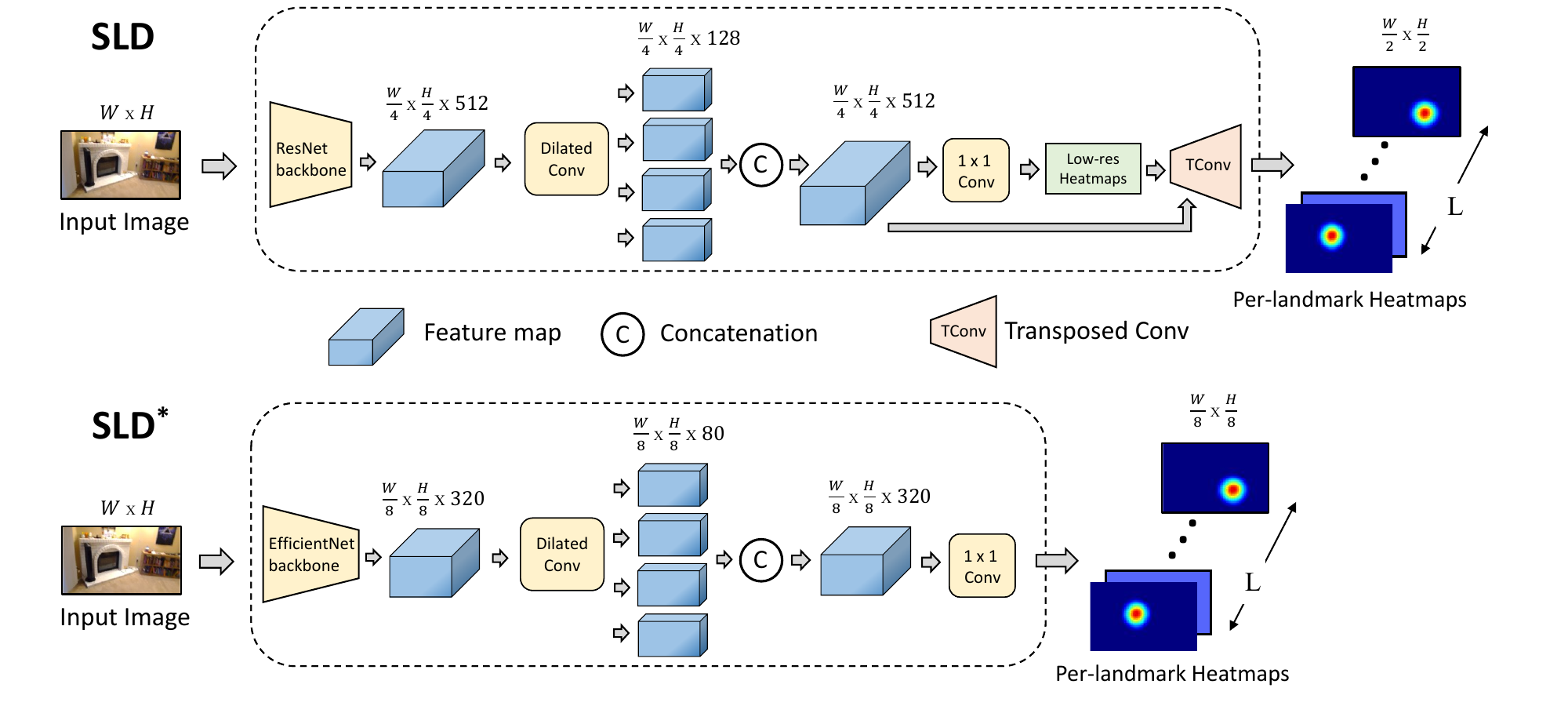}
  \caption{[Top] The original SLD architecture ~\cite{Do2022}. [Bottom] An illustration of the proposed SLD\textsuperscript{$\ast$} architecture (see text for details).}
  \label{fig:sldstar}
  \vspace{-2mm}
\end{figure*}

\customparagraph{Landmark Selection.} 3D scene points with discriminative appearance that are associated with permanent scene structures can serve as good scene landmarks. Do \etal~\cite{Do2022} proposed a greedy method to select landmarks, given SfM camera poses, 3D points and the associated 2D image observations. Their method heuristically selects groups of 3D points that are well distributed within the  scene. We use the same method, but experiment with up to 1500 landmarks, in contrast to the use of 200--400 points in prior work~\cite{Do2022}.

\customparagraph{Model.} The SLD architecture is fully convolutional and inspired by existing neural architectures for keypoint prediction in images using heatmaps. Do \etal~\cite{Do2022} implemented SLD using both ResNet-18~\cite{He_2016_resnet} and EfficientNet~\cite{TanEfficientNet20} backbones. The features from the backbone network are then passed into a dilated convolution layer~\cite{Yu2015MultiScaleCA} followed by a 1$\times$1 convolution layer to produce low-resolution heatmaps. Finally, the heatmaps are upsampled using a transposed convolution layer. The architecture is illustrated in the upper part of Figure~\ref{fig:sldstar}. In contrast to SLD, the NBE network uses fully connected layers after a ResNet-18 backbone that output the final bearing predictions. SLD and NBE models were trained on the same scene and the authors proposed running inference using both models on every query image.

\customparagraph{Training.} The SLD and NBE architectures are trained using ground truth 2D landmark detections (and 3D bearing vectors) derived from associated camera poses in the training data. Training SLD also requires knowledge about which images each landmark is visible in. The visibilities are recovered from 2D data association of SfM 3D points in the training images. SLD is then trained using mean squared loss with respect to the ground truth heatmaps, while NBE is trained with a robust angular loss.

\customparagraph{Datasets and Metrics.}
SLD and NBE was evaluated on \textsc{Indoor-6}~\cite{Do2022}, a challenging indoor localization dataset with six scenes, where images captured over multiple days have strong lighting changes. Pseudo ground truth (pGT) camera poses were recovered with COLMAP~\cite{Schoenberger2016Structure}. Given, camera pose estimates, the standard rotational error $\Delta R$ and position error $\Delta t$ is computed as follows.
\[
\Delta R = \arccos{ \tfrac{ \text{Tr}(\mathbf{R}^{\top} \mathbf{\hat{R}}) - 1 }{2} }, \ \ \ \Delta t = \Vert \mathbf{R}^{\top} \mathbf{t} - \mathbf{\hat{R}}^{\top} \mathbf{\hat{t}} \Vert_2.\\
\]
given ($\mathbf{R}$, $\mathbf{t}$) and ($\mathbf{\hat{R}}$, $\mathbf{\hat{t}}$), the estimated and ground truth poses respectively.
The final metric is recall at 5cm/5$^{\circ}$, the fraction of test images where $\Delta R \leq 5^\circ$ and $\Delta t \leq 5$cm.

\subsection{SLD\textsuperscript{$\ast$} Architecture}

In this section, we introduce SLD\textsuperscript{$\ast$}, a more compact and memory efficient architecture, and an improved pose solver. Next, we highlight the four key differences with SLD+NBE. Figure~\ref{fig:sldstar} compares the SLD and 
SLD\textsuperscript{$\ast$} architectures.

\begin{figure*}[t]
\centering
\includegraphics[width=0.98\linewidth]{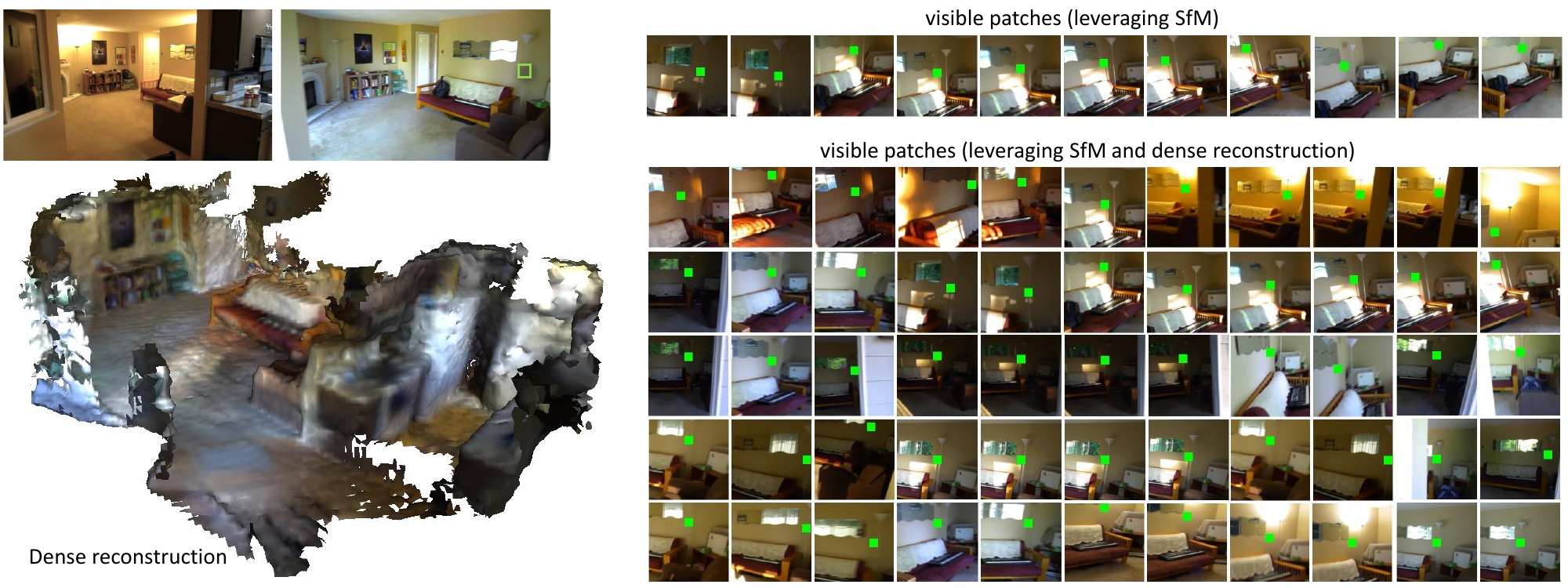}
  \caption{\textbf{Better Visibility Estimation}. [Left] Two images from scene1 in the \textsc{Indoor-6} dataset taken at different times of day and a rendering of the dense 3D mesh reconstruction of the scene. [Right] On the top right, we show a single row of patches depicting a scene landmark (indicated by the green square) in different images where the landmark was found to be visible. The original method leveraged data association from only structure from motion. On the lower right, we show patches for the same landmark based the proposed visibility estimation approach that also uses the dense mesh reconstruction (see text for details). The high appearance diversity in the observed patches under varying illumination makes the trained landmark detector more robust.}
  \label{fig:scene1MeshAndPatches}
\end{figure*}

\customparagraph{NBE not used.} Do \etal ~\cite{Do2022} proposed using NBE to directly regress bearing vectors of the landmarks even when they were not visible in the image. These bearing predictions were complementary to SLD's heatmap detections. As SLD's typical budget of landmarks is quite small, sometimes enough landmarks are not visible in a test image.
However, the steps to merge the two sets of predictions is adhoc. SLD\textsuperscript{$\ast$} does not use NBE, as it uses a larger landmark budget to directly address the underlying issue.

\customparagraph{Absence of an upsampling layer.} SLD first predicts a set of low-resolution heatmaps and then spatially upsamples them using transposed convolutions to produce the final heatmaps. In contrast, SLD\textsuperscript{$\ast$} directly predicts the output heatmaps using 1$\times$1 convolution without any spatial upsampling. Without the upsampling layer, SLD\textsuperscript{$\ast$} has fewer parameters to learn and has a smaller memory footprint. Yet, this change does not adversely affect the accuracy of landmark prediction in our experience. This is because, for each detected landmark, the associated 2D position is estimated by computing a weighted mean of all the heatmap samples from a 17$\times$17 patch centered at the location of the peak in each heatmap. We observe that the weighted averaging step provided sufficient sub-pixel precision in the 2D landmark coordinates and thus predicting heatmaps at a high output resolution appears to be unnecessary.

\customparagraph{Memory footprint reduction.} Do \etal~\cite{Do2022} experimented with both ResNet-18~\cite{He_2016_resnet} and EfficientNet~\cite{TanEfficientNet20} backbones. In our implementation, we focus only on EfficientNet, as we aim to reduce the storage size and the memory footprint of the architecture. Furthermore, we use fewer feature map channels and more aggressive downsampling than SLD. SLD\textsuperscript{$\ast$} has 320 channels unlike SLD which has 512 channels. SLD\textsuperscript{$\ast$}'s feature maps have 8$\times$ downsampling in contrast to SLD, where the downsampling factor is 4$\times$.

\customparagraph{Weighted pose estimation.}
We implemented a weighted pose estimation scheme using weights derived from the heatmap values associated with SLD\textsuperscript{$\ast$}'s output predictions. Denoting peak heatmap values per detection as $v$, we first prune detections for which $v \le 0.3$. Next, we compute a per-landmark weight $w = v^e$ where $e$ is a parameter. We propose using the weights $w$ in two different steps. First, for PROSAC~\cite{Prosac2005} (RANSAC variant) used for robust estimation and also as weights during the PnP pose optimization. 
\subsection{Landmark Visibility Estimation}
In this section, we discuss a limitation of how training data is generated for SLD~\cite{Do2022} and propose methods for addressing the limitation. While SfM pipelines such as COLMAP~\cite{Schoenberger2016Structure} can produce 3D points with accurate 2D data association in multiple images, they often fail to detect all the potential observations (true positives) of the point. This can happen when the illumination varies dramatically. To alleviate this issue, Do \etal~\cite{Do2022} proposed an ad-hoc augmentation strategy where they assumed that a landmark is visible in images whose camera poses estimated by SfM are nearby to the pose of images where the point is known to be visible. However, this strategy can corrupt the training data with outliers (false positives) by including views where the landmark is occluded. We propose to mitigate this issue using geometry and explicit occlusion reasoning.
\begin{table*}[t]
\small
\begin{adjustbox}{max width=\textwidth,center}
\renewcommand\arraystretch{1.1}

\parbox{.32\linewidth}{
\centering
\begin{tabular}{lcccc}
\hline
& \multicolumn{4}{c}{Num. Landmarks}\\
& 100 & 200 & 300 & 400\\
\hline
scene1  & 34.7 & 39.8 & 41.8 & 17.6\\
scene2a & 31.5 & 46.3 & 45.9 & 28.8\\
scene3  & 34.3 & 43.2 & 55.2 & 42.5\\
scene4a & 46.2 & 63.3 & 65.8 & 42.4\\
scene5  & 28.5 & 31.4 & 35.1 & 29.7\\
scene6  & 43.3 & 58.2 & 56.4 & 40.3\\
\hline
avg. & 36.4 & 47.0 & 50.0 & 33.6\\
\hline \\
\multicolumn{5}{c}{(a)~\textbf{Pose recall at 5cm/5$^{\circ}$} (in \%) $\uparrow$}\\
\end{tabular}
}

\hfill

\parbox{.32\linewidth}{
\centering
\begin{tabular}{lcccc}
\hline
& \multicolumn{4}{c}{Num. Landmarks}\\
& 100 & 200 & 300 & 400\\
\hline
scene1  & 0.29 & 0.39 & 0.41 & 0.53\\
scene2a & 0.24 & 0.25 & 0.34 & 0.43\\
scene3  & 0.27 & 0.29 & 0.36 & 0.45\\
scene4a & 0.23 & 0.28 & 0.29 & 0.44\\
scene5  & 0.25 & 0.39 & 0.40 & 0.53\\
scene6  & 0.25 & 0.28 & 0.30 & 0.46\\
\hline
avg. & 0.26 & 0.31 & 0.35 & 0.47\\
\hline \\
\multicolumn{5}{c}{(b)~\textbf{angular error (in deg.)} $\downarrow$}\\
\end{tabular}
}

\hfill

\parbox{.32\linewidth}{
\centering
\begin{tabular}{lcccc}
\hline
& \multicolumn{4}{c}{Num. Landmarks}\\
& 100 & 200 & 300 & 400\\
\hline
scene1  & 0.29 & 0.34 & 0.35 & 0.49\\
scene2a & 0.24 & 0.26 & 0.31 & 0.37\\
scene3  & 0.27 & 0.29 & 0.31 & 0.39\\
scene4a & 0.23 & 0.26 & 0.28 & 0.41\\
scene5  & 0.25 & 0.32 & 0.35 & 0.41\\
scene6  & 0.25 & 0.29 & 0.34 & 0.45\\
\hline
avg. & 0.26 & 0.29 & 0.32 & 0.42\\
\hline \\
\multicolumn{5}{c}{(c)~\textbf{angular error (in deg.) (first 100)} $\downarrow$}\\
\end{tabular}
}
\end{adjustbox}
\caption{\textbf{Analyzing Model Capacity:} (a) The table reports averge camera pose estimation accuracy according to the 5cm/5$^{\circ}$ recall metric for four SLD\textsuperscript{$\ast$} models trained with 100, 200, 300 and 400 landmarks respectively for all the scenes in \textsc{Indoor-6}. (b) The median angular error in degrees for the same four models averaged across the six scenes. The median is computed over the set of all 2D SLD detections obtained using the trained models on all the test images. (c) In our implementation, the elements in the selected set of landmarks are stored in the order they were selected. Therefore, the fist 100 landmarks in the ordered sets for the models trained on 100, 200, 300 and 400 landmarks are identical. The median errors for the first 100 landmarks averaged on all scenes, are reported in the table.}
\label{tab:analysis}
\end{table*}

\customparagraph{Dense Reconstruction.}
We reconstruct a dense 3D mesh for each scene as follows.
First, dense monocular depth maps for all map images are estimated using the dense depth vision transformer~\cite{ranftl2020towards}. The dense 3D point clouds from these depth maps are then robustly registered to the sparse SfM 3D point cloud (which is computed by COLMAP~\cite{Schoenberger2016Structure}). The registration involves robustly estimating an affine transformation from 3D point-to-point matches. We first prune 3D points observed in less than 50 images and remove images which did not observe a sufficient number of 3D points. We also check residuals after aligning the depth maps to the SfM points and prune out images for which the mean depth residual exceeded 5cm. Finally, we use truncated signed distance function based depth-map fusion and isosurface extraction to compute the mesh. Figure~\ref{fig:scene1MeshAndPatches} shows the reconstructed mesh for scene1.

\customparagraph{Occlusion Reasoning.}
For every pair of a selected landmark $\mathbf{p}$ and an image $\mathcal{I}$ with its pose $\mathbf{T}_{\mathcal{I}}$ and the estimated dense depth $d_{\mathcal{I}}$ we determine whether the landmark is visible in the image by checking the following conditions:
\begin{itemize}
    \item The 3D point $\mathbf{p}$ is in front of the camera for $\mathcal{I}$ (i.e., $(\mathbf{T}_{\mathcal{I}}\mathbf{p}_l)_z > 0$) and the point projects within the image (i.e., $\Pi(\mathbf{T}_{\mathcal{I}}\mathbf{p}_l)) \in \mathcal{R}(\mathcal{I})$ where $\Pi(.)$ is the 2D projection operator and $\mathcal{R}(.)$ denotes the image extent.
    \item The depth of the 2D projected point is not too far from the depth at that pixel, computed using the reconstructed mesh, i.e., $d_{\mathcal{I}}(\Pi(\mathbf{T}_{\mathcal{I}}\mathbf{p})) \approx (\mathbf{T}_{\mathcal{I}}\mathbf{p})_z$.
    \item The surface normal of the 2D projected point is not too far from the normal vector estimated using the reconstructed mesh, i.e, $\nabla d_{\mathcal{I}}(\Pi(\mathbf{T}_{\mathcal{I}}\mathbf{p})) \approx \nabla(\mathbf{T}_{\mathcal{I}}\mathbf{p})_z$
\end{itemize}

\subsection{Landmark Partitioning For Scalability}
In this section, we discuss what prevents SLD from accurately scaling to a large number of landmarks and present a simple solution that does not add  computational overhead.

\customparagraph{Insufficient Capacity.} Do \etal~\cite{Do2022} evaluated SLD (with ResNet-18) models with 200, 300 and 400 landmarks per scene on \textsc{Indoor-6} and reported that 300 landmarks worked best. When evaluating SLD\textsuperscript{$\ast$} with different number of landmarks, we observed that accuracy increases from 100 to 300 but falls with 400 landmarks (see the recall at 5cm/5$^{\circ}$ metrics in Table ~\ref{tab:analysis}(a)). It is worth noting that the smaller sets of landmarks are strictly contained within the larger landmark sets. The results imply that insufficient model capacity in the network could be hurting accuracy.

To confirm our hypothesis, we analyzed the angular errors of the predicted 2D landmarks from the SLD\textsuperscript{$\ast$} models trained on 100, 200, 300 and 400 landmarks. The median angular errors reported in Table~\ref{tab:analysis}(b) increased as the number of landmarks increased. The angular errors depend only on the network, and are not affected by pose estimation or other factors. We also analyzed the angular error of the first 100 landmarks (defined with respect to an ordering defined by landmark ids) for the four SLD\textsuperscript{$\ast$} models trained on 100, 200, 300 and 400 landmarks. Since the first 100 landmarks are identical in all four cases, comparing the median errors on these 100 points in the four models is the best way to compare them. Indeed as Table~\ref{tab:analysis}(c) shows, the predictions for the first 100 landmarks get worse as the model is trained for 200, 300 and 400 landmarks. This confirms our hypothesis that the models have insufficient capacity.

\begin{figure*}
\captionsetup[subfigure]{labelformat=empty}
\centering
\begin{subfigure}{0.41\columnwidth}
\includegraphics[width=\textwidth]{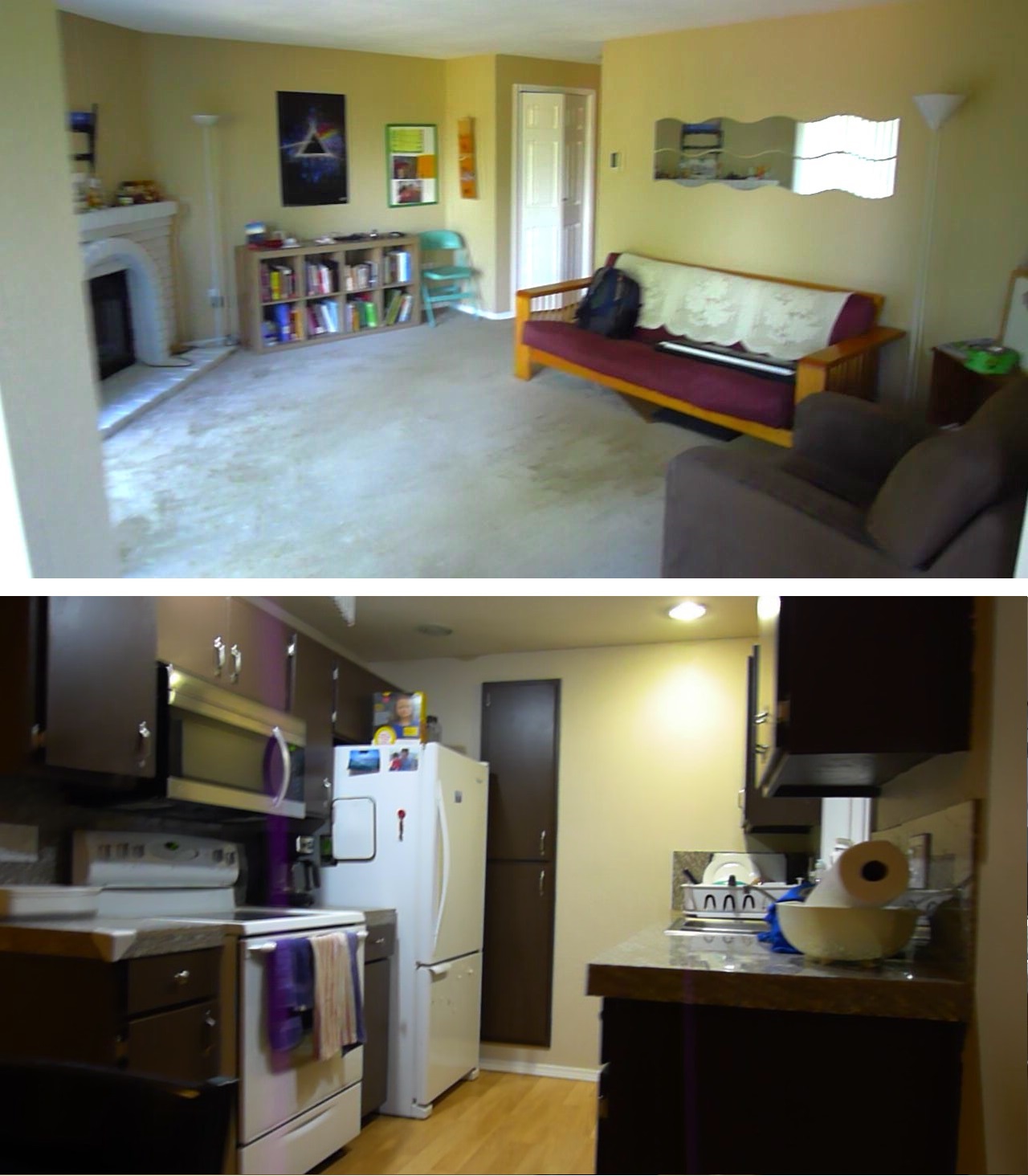}
\caption{Scene1 images}
\end{subfigure}
\begin{subfigure}{0.5\columnwidth}
\includegraphics[width=\textwidth]{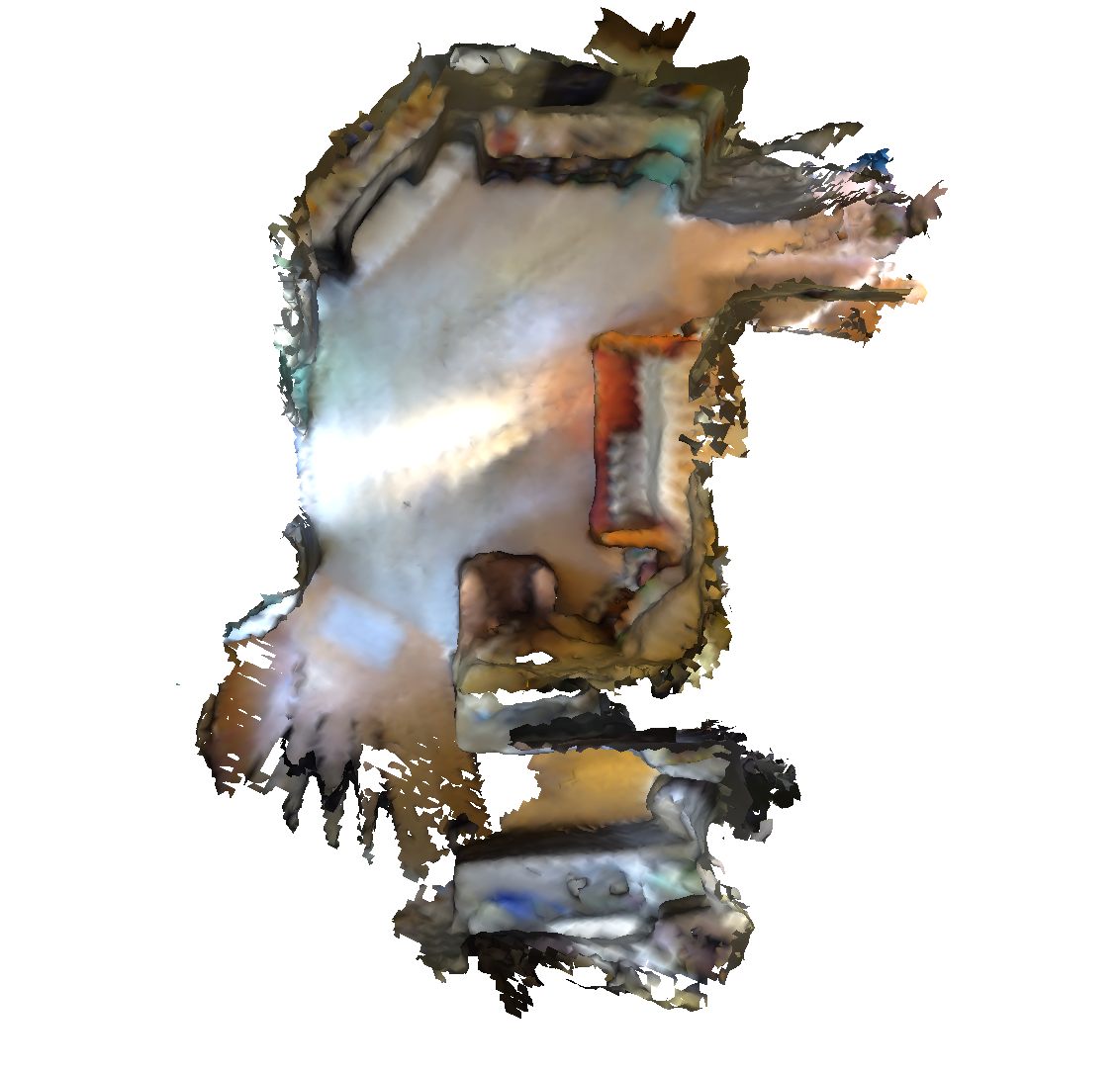}
\caption{Scene1 mesh}
\end{subfigure}
\begin{subfigure}{0.53\columnwidth}
\includegraphics[width=\textwidth]{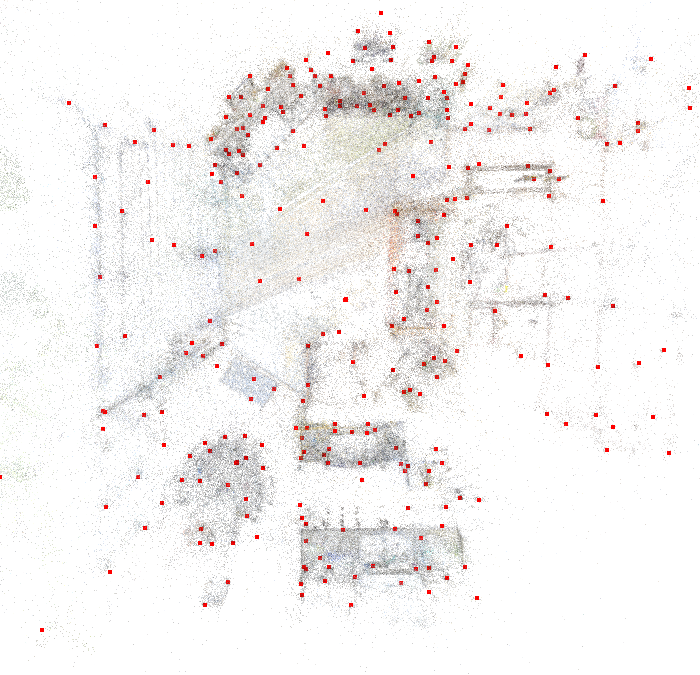}
\caption{300 landmarks}
\end{subfigure}
\begin{subfigure}{0.53\columnwidth}
\includegraphics[width=\textwidth]{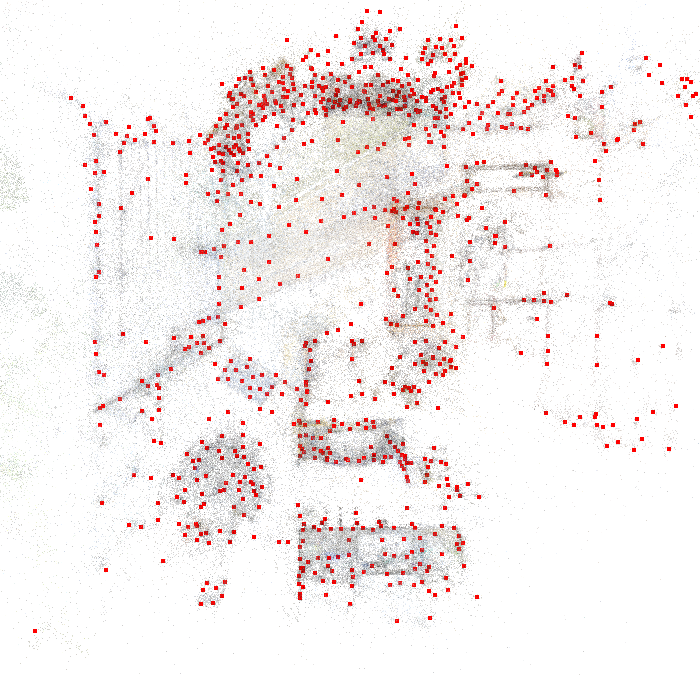}
\caption{1000 landmarks}
\end{subfigure}
\caption{The top view of the mesh and 3D SfM point cloud from scene1, shown with the overlaid scene landmarks (red points). The sets of 300 and 1000 landmarks respectively are both computed by the existing selection method~\cite{Do2022}. The image on the right shows that a higher number of landmarks provides denser scene coverage. We show later that it leads to an improvement in camera pose accuracy.}
\label{fig:scene1_landmarks}
\end{figure*}

\customparagraph{Training network ensembles.}
Instead of modifying the architecture, we address the insufficient capacity issue by choosing a divide and conquer strategy for scaling to a higher number of landmarks. We propose to simply partition the set of landmarks into non-overlapping subsets where the subsets are relatively small and their size is selected by keeping the typical capacity of the SLD\textsuperscript{$\ast$} architecture under consideration. Then, we independently train multiple identical networks, one for each subset. We refer to the networks together as an ensemble. The networks in the ensemble can be trained independently and each is aware only of its own associated subset of landmarks. Training a SLD\textsuperscript{$\ast$} ensemble is thus trivially parallelizable. 

\customparagraph{Parallel vs. Sequential Inference.} At test time, there are two ways to run inference using the ensemble. When GPU memory is abundant, all SLD\textsuperscript{$\ast$} networks could be initialized in GPU memory, allowing parallel inference on multiple networks. Despite having multiple networks, the total memory footprint can still be quite reasonable as each SLD\textsuperscript{$\ast$} network is quite memory efficient ($<$ 0.99 GB). In this setting, inference can be extremely fast and real-time processing is quite viable.
However, on GPUs with smaller memory budgets, inference must be done sequentially. Even though, the processing time grows, localization can still run at 3--5 images/sec for practical ensemble sizes. In this paper, all reported timings are for the sequential inference setting.

\customparagraph{Partitioning Criteria.}
We compare four different criteria for partitioning the landmark set -- (1) Default: sorting landmarks by the saliency score and then splitting the sorted list into equal sized partitions; (2) Random: randomly assigning landmarks to partitions; (3) Spatial clustering: grouping landmarks by k-means clustering and then rebalancing points in adjoining clusters to get equal sized partitions; (4) Farthest-point sampling: 
iteratively selecting the point farthest from the points already in existing partitions and adding it to the best partition until all partitioned reached the specified size.
We compared the four criteria using 1000 landmarks and 8 partitions and found that the recall at 5cm/5$^{\circ}$ pose metric was similar (within 1-2 \% points) in the four cases. We conclude that the partitioning criteria is not crucial on the dataset and thus used the default strategy thereafter. However, when a coarse location prior is available in large scenes, clustering-based partitioning can improve computational efficiency by enabling locality-based pruning of redundant inference passes.

\begin{table}
\centering
\small
\begin{adjustbox}{max width=\columnwidth,center}
\renewcommand\arraystretch{1.2}
\begin{tabular}{l|c|c|c|c|c|c|c}
& 200$\times$1 & 300$\times$1 & 100$\times$3 & 100$\times$4 & 125$\times$6 & 125$\times$8 & 125$\times$12\\
\hline
R @ 5cm/5$^{\circ}$ $\uparrow$ & 46.0 & 50.8& 61.1 & 63.0 & 66.6 & \textbf{70.1}& 69.1\\
Time (sec.) $\downarrow$& \textbf{0.05} & 0.11 & 0.16 & 0.19 & 0.23 & 0.3 & 0.5 \\
Size (MB) $\downarrow$ & \textbf{15} & \textbf{15} & 45 & 60 & 90 & 120 & 180\\
\hline
\end{tabular}
\end{adjustbox}
\caption{\textbf{Ablation study.} 
Recall at 5cm/5$^{\circ}$ for a$\times$b ensembles where a is the number of landmarks in each subset and b is the number of networks in the ensemble. The 125$\times$8 ensemble has the best performance. As expected, ensembles containing more networks and dealing with more scene landmarks have slightly higher storage requirements and running times.}
\label{table:ensemble}
\end{table}

\begin{table}
\centering
\small
\begin{adjustbox}{max width=\columnwidth,center}
\renewcommand\arraystretch{1.2}
\begin{tabular}{l|c|c|c|c|c}
& \sout{w}& w = v & w = v$\sqrt{v}$& w = v$^2$ & w = v$^2\sqrt{v}$\\
\hline
R @ 5cm/5$^{\circ}$ $\uparrow$ & 68.0\% & 68.4\% & 69.4\% &\textbf{70.1}\% & 69.6\\
\hline
\end{tabular}
\end{adjustbox}
\caption{\textbf{Evaluating weighted pose estimation schemes.} Recall at 5cm/5$^{\circ}$ of a 1000 landmark
SLD\textsuperscript{$\ast$} ensemble on \textsc{Indoor-6} for non-weighted (\sout{v}) and weighted pose estimates. Four schemes for deriving the 
weights (w) from heatmap values (v) are compared.}
\label{table:abla}
\end{table} 
\section{Experimental Results}

\begin{table*}[t]
\normalsize
\begin{adjustbox}{max width=\textwidth,center}
\renewcommand\arraystretch{1.5}
\begin{tabular}{l|l|c|cc|c|c|cc|cc|c}
                             & Scene    & DSAC*     & NBE+SLD  & SLD    & SegLoc & SLD\textsuperscript{$\ast$} & hloc-l${}_{\scriptsize{1000}}$ & hloc-l${}_{\scriptsize{3000}}$ & hloc-A & hloc-B & SLD\textsuperscript{$\ast$}\\
&            & \cite{brachmann2021dsacstar}& \cite{Do2022} & \cite{Do2022} & \cite{Pietrantoni_2023_CVPR} & \textbf{ours} & \cite{Do2022} & \cite{Do2022} & \cite{Do2022,sarlin2019} & \cite{sarlin2019} & \textbf{ours}\\
\hline
\#landmarks                  &          &  n/a      & 300      & 300    & n/a    & 300        & 1000       & 3000       & n/a         & n/a             & 1000 \\
\hline
R@5cm/5$^{\circ}$ $\uparrow$ & scene1   & 18.7      & 38.4     & 35.0   & 51.0  & 47.2       & 33.3       & 48.1       & 64.8        & \textbf{70.5}   & \blue{68.5}\\
                             & scene2a  & 28.0      & --       & 34.6   & 56.4  & 48.2       & 12.5       & 17.1       & 51.4        & \blue{52.1}     & \textbf{62.6}\\
                             & scene3   & 19.7      & 53.0     & 50.8   & 41.8  & 56.2       & 48.3       & 61.9       & \blue{81.0} & \textbf{86.0}   & 76.2\\
                             & scene4a  & 60.8      & --       & 56.3   & 33.8  & 67.7       & 34.8       & 39.2       & 69.0        & \blue{75.3}     & \textbf{77.2}\\
                             & scene5   & 10.6      & 40.0     & 43.6   & 43.1  & 33.7       & 21.9       & 31.1       & 42.7        & \textbf{58.0}   & \blue{57.8}\\
                             & scene6   & 44.3      & 50.5     & 48.9   & 34.5  & 52.0       & 47.4       & 59.1       & \blue{79.9} & \textbf{86.7}   & 78.0\\
\hline
R@5cm/5$^{\circ}$ $\uparrow$ & avg.     & 30.4      & 45.5     & 44.9   & 43.4  & 50.8       & 33.0       & 42.8       & 64.8        & \textbf{71.4}   & \blue{70.1}\\
\hline
 Size (GB) $\downarrow$      &          & 0.027     & 0.135    & 0.020  & 0.161 & 0.015      & 0.17--0.21 & 0.2--0.5   & 0.7--2.4    & 0.7--2.4        & 0.120 \\
 \hline
 Mem. (GB) $\downarrow$      &          & 0.85      & 1.35     & 1.2    & --    & 0.99       & 1.3        & 1.3        & 1.3         & 1.3             & 0.99 \\
 \hline
\end{tabular}
\end{adjustbox}
\caption{\textbf{Quantitative Evaluation on \textsc{Indoor-6}.} We report the recall at 5cm/5$^\circ$  (in \%), storage used (Size), and in-memory footprint (Mem.) of several methods.
For the SLD~\cite{Do2022} baseline, we report previously published results in the column NBE+SLD and results from the public EfficientNet-based code in the SLD column. For hloc~\cite{sarlin2019}, we first present published results in Do ~\etal \cite{Do2022} in the column hloc-A, and then, the best results we obtained using hloc's public codebase in the column hloc-B. Finally, we present results for SLD\textsuperscript{$\ast$} (denoted "ours") with 300 and 1000 landmarks respectively. The best method (per row) is highlighted in bold and the second-best in blue.}
\label{table:recall}
\end{table*}

\begin{figure*}
\centering
\includegraphics[width=0.95\linewidth]{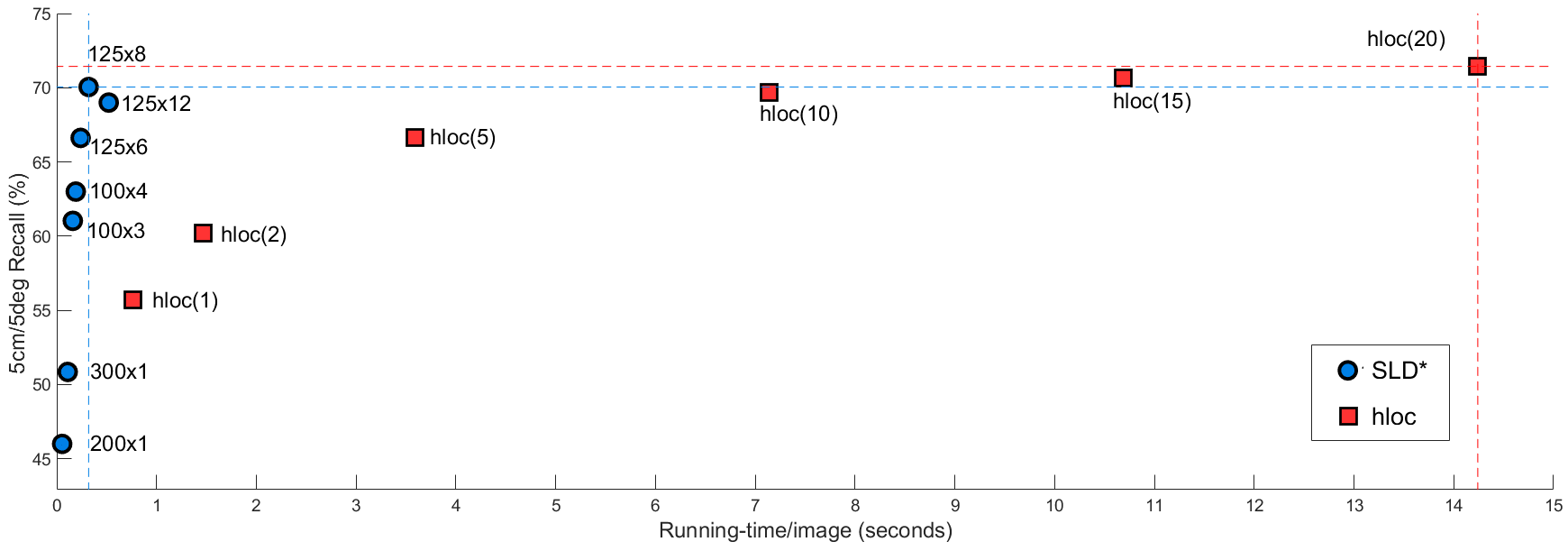}
\caption{\textbf{Accuracy/speed tradeoff of SLD\textsuperscript{$\ast$} and hloc.} 
The plot shows how hloc's performance varies with the number of matched image pairs. Tthe number of pairs were set to 1, 2, 5, 10, 15 and 20 respectively, as denoted by the text labels).
hloc's best accuracy was 71.4\% with 20 image pairs for which the timing was 14.2 seconds/image. Similarly, seven SLD\textsuperscript{$\ast$} configurations were evaluated. 
The text label a $\times$ b next to the blue dots indicate the SLD\textsuperscript{$\ast$} configuration, where a is the number of landmarks in each partition and b represents the number of partitions. SLD\textsuperscript{$\ast$}'s best result was 70.1\% using 125 $\times$ 8 = 1000 landmarks with a running time of 0.3 seconds/image. The plot shows that accuracy wise, SLD\textsuperscript{$\ast$}'s best configuration is competitive with hloc but more than 40X faster.}
\label{fig:scatterplot}
\end{figure*}

In this section, we report ablation studies and a quantitative comparison of SLD\textsuperscript{$\ast$} and other methods on \textsc{Indoor-6}. We then study in detail, the accuracy and speed tradeoff of SLD\textsuperscript{$\ast$} and hloc~\cite{sarlin2019}. Finally, we present visual examples to show the benefit of using a larger number of landmarks. 

\customparagraph{Ablation: Ensemble Size.}
Table~\ref{table:ensemble} shows 5cm/5$^{\circ}$ recall for a variety of ensemble sizes that we have evaluated. We empirically found that 125$\times$8 (8 networks with 125 landmarks each) works the best on \textsc{Indoor-6}. We also report how storage and running times increase proportional to the ensemble size and the total number of landmarks.

\customparagraph{Ablation: Weighted Pose Estimation.}
Table~\ref{table:abla} reports 5cm/5$^{\circ}$ recall for non-weighted and weighted pose estimation using a 125$\times$8 SLD\textsuperscript{$\ast$} ensemble. For the weighted case, the effect of setting values of the parameter $e$ to 1, 1.5, 2 and 2.5 is reported. The setting $e$ = 2 gave the best results and was used in all the other experiments.

\customparagraph{Quantitative Evaluation.}
In Table~\ref{table:recall}, we compare recall at 5cm/5$^{\circ}$ for several methods. For \textbf{DSAC*}~\cite{brachmann2021dsacstar}, \textbf{SegLoc}~\cite{Pietrantoni_2023_CVPR}, \textbf{NBE+SLD}~\cite{Do2022}
we present results reported in prior work~\cite{Do2022,Pietrantoni_2023_CVPR}. The \textbf{SLD} column in the table shows the results of the public EfficientNet-based SLD implementation. The table also includes results of hloc~\cite{sarlin2019}. Previously reported results are shown in column hloc-A whereas our results obtained with hloc's public implementation are shown in column hloc-B. Finally, hloc-lite (hloc-l) results from prior work~\cite{Do2022} are also included.

\begin{figure*}[t]
\captionsetup[subfigure]{labelformat=empty}
\centering
\footnotesize
\begin{subfigure}{0.49\columnwidth}
\includegraphics[width=\textwidth]{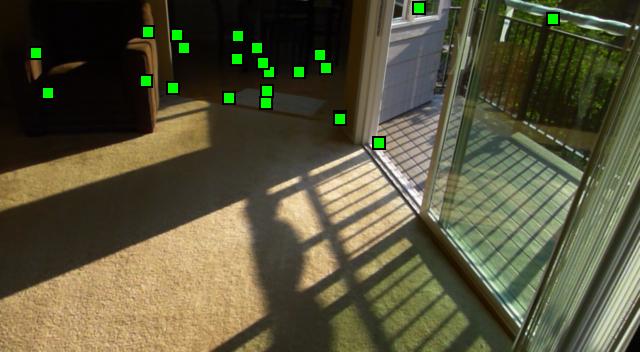}
\caption{[L=300] $\Delta R$~=~$0.94^\circ$, $\Delta t$~=~$8$cm}
\end{subfigure}
\hspace{1pt}
\begin{subfigure}{0.49\columnwidth}
\includegraphics[width=\textwidth]{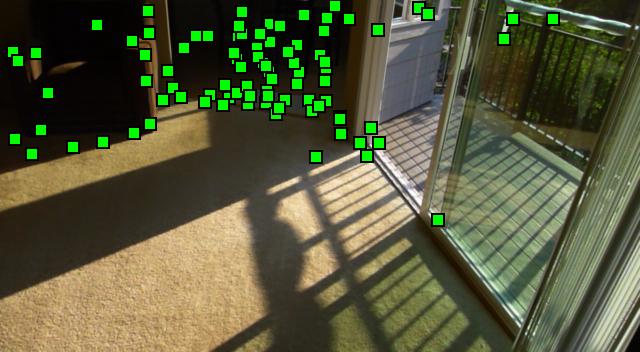}
\caption{[\blue{L=1000}] $\Delta R$~=~$0.25^\circ$, $\Delta t$~=~$4$cm}
\end{subfigure}
\hspace{4mm}
\begin{subfigure}{0.49\columnwidth}
\includegraphics[width=\textwidth]{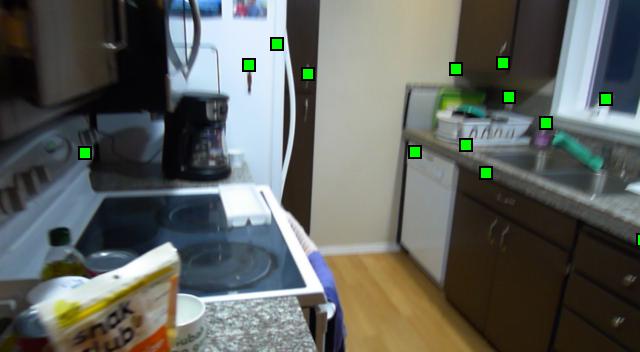}
\caption{[L=300] $\Delta R$~=~$1.73^\circ$, $\Delta t$~=~$10$cm}
\end{subfigure}
\hspace{1pt}
\begin{subfigure}{0.49\columnwidth}
\includegraphics[width=\textwidth]{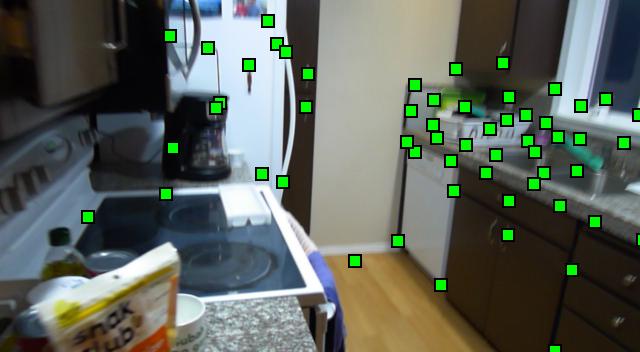}
\caption{[\blue{L=1000}] $\Delta R$~=~$0.18^\circ$, $\Delta t$~=~$2$cm}
\end{subfigure}\\
(a) \hspace{8.4cm} (b)\\
\vspace{3mm}
\begin{subfigure}{0.49\columnwidth}
\includegraphics[width=\textwidth]{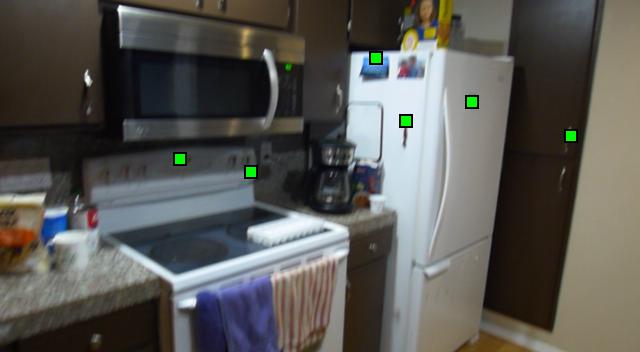}
\caption{[L=300] $\Delta R$~=~$1.46^\circ$, $\Delta t$~=~$5$cm}
\end{subfigure}
\hspace{1pt}
\begin{subfigure}{0.49\columnwidth}
\includegraphics[width=\textwidth]{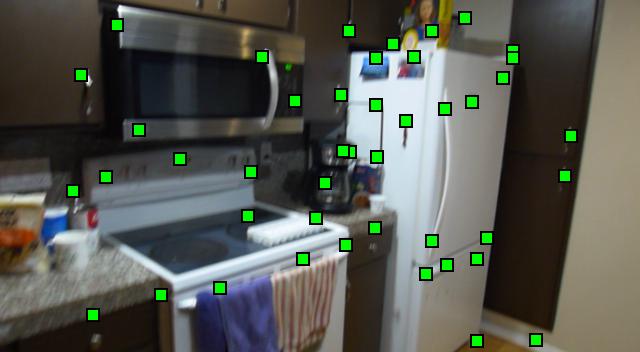}
\caption{[\blue{L=1000}] $\Delta R$~=~$0.29^\circ$, $\Delta t$~=~$1$cm}
\end{subfigure}
\hspace{4mm}
\begin{subfigure}{0.49\columnwidth}
\includegraphics[width=\textwidth]{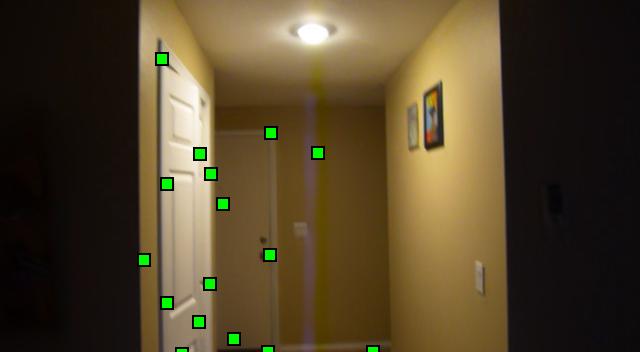}
\caption{[L=300] $\Delta R$~=~$0.81^\circ$, $\Delta t$~=~$12$cm}
\end{subfigure}
\hspace{1pt}
\begin{subfigure}{0.49\columnwidth}
\includegraphics[width=\textwidth]{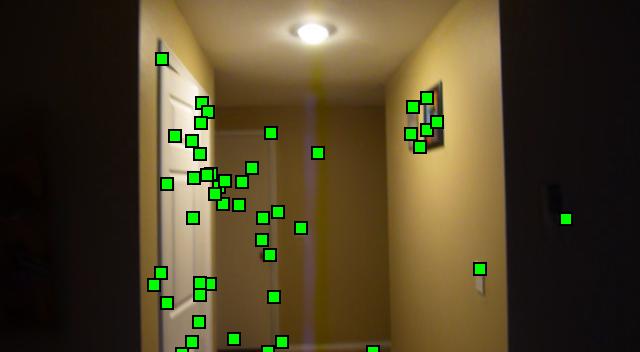}
\caption{[\blue{L=1000}] $\Delta R$~=~$0.77^\circ$, $\Delta t$~=~$4$cm}
\end{subfigure}
\\
(c) \hspace{8.4cm} (d)\\
\vspace{3mm}
\begin{subfigure}{0.49\columnwidth}
\includegraphics[width=\textwidth]{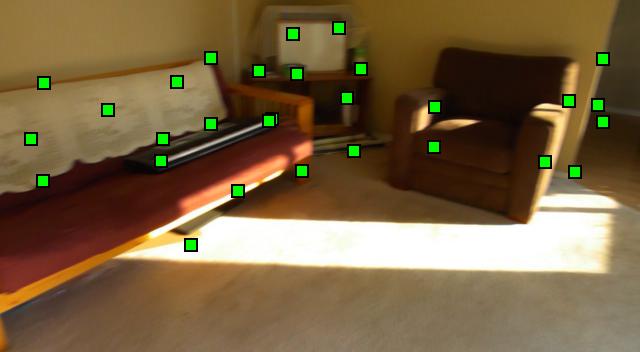}
\caption{[L=300] $\Delta R$~=~$0.69^\circ$, $\Delta t$~=~$5$cm}
\end{subfigure}
\hspace{1pt}
\begin{subfigure}{0.49\columnwidth}
\includegraphics[width=\textwidth]{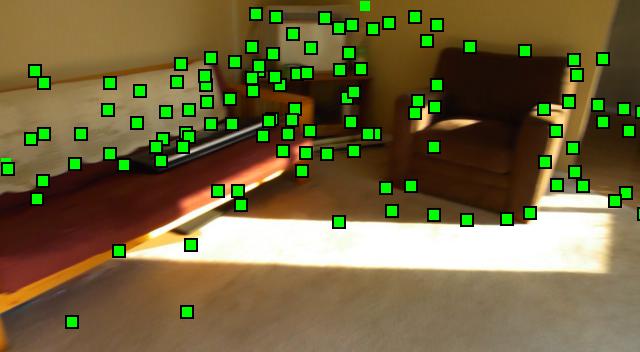}
\caption{[\blue{L=1000}] $\Delta R$~=~$0.28^\circ$, $\Delta t$~=~$2$cm}
\end{subfigure}
\hspace{4mm}
\begin{subfigure}{0.49\columnwidth}
\includegraphics[width=\textwidth]{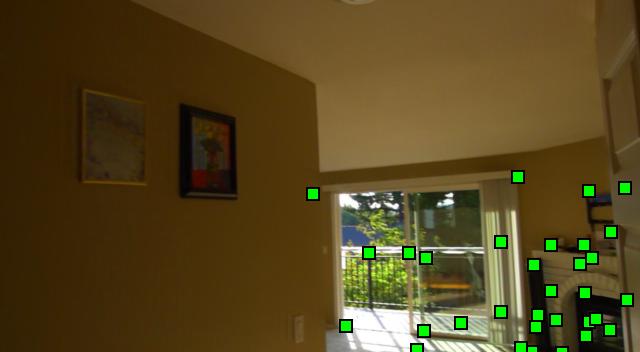}
\caption{[L=300] $\Delta R$~=~$2.15^\circ$, $\Delta t$~=~$38$cm}
\end{subfigure}
\hspace{1pt}
\begin{subfigure}{0.49\columnwidth}
\includegraphics[width=\textwidth]{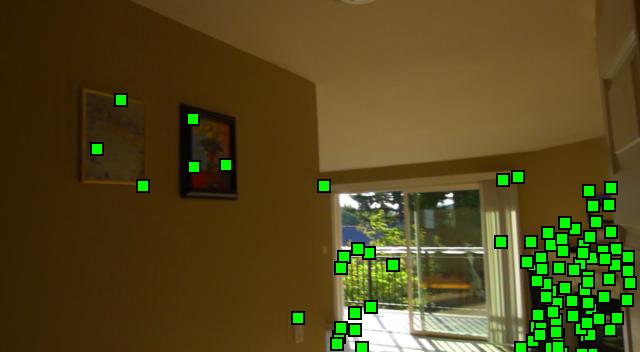}
\caption{[\blue{L=1000}] $\Delta R$~=~$0.21^\circ$, $\Delta t$~=~$7$cm}
\end{subfigure}
\\
(e) \hspace{8.4cm} (f)\\
\vspace{3mm}
\begin{subfigure}{0.49\columnwidth}
\includegraphics[width=\textwidth]{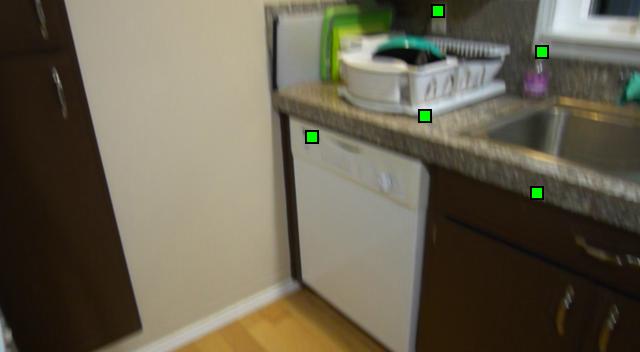}
\caption{[L=300] $\Delta R$~=~$2.47^\circ$, $\Delta t$~=~$6$cm}
\end{subfigure}
\hspace{1pt}
\begin{subfigure}{0.49\columnwidth}
\includegraphics[width=\textwidth]{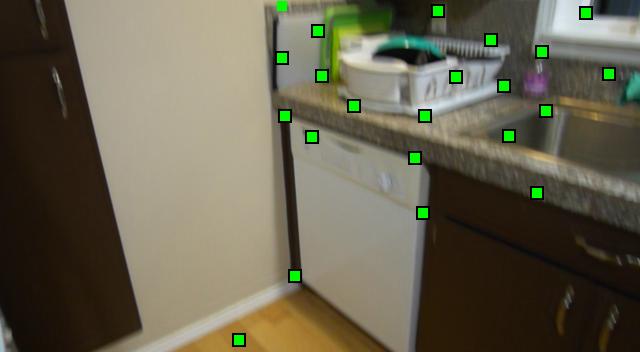}
\caption{[\blue{L=1000}] $\Delta R$~=~$0.76^\circ$, $\Delta t$~=~$3$cm}
\end{subfigure}
\hspace{4mm}
\begin{subfigure}{0.49\columnwidth}
\includegraphics[width=\textwidth]{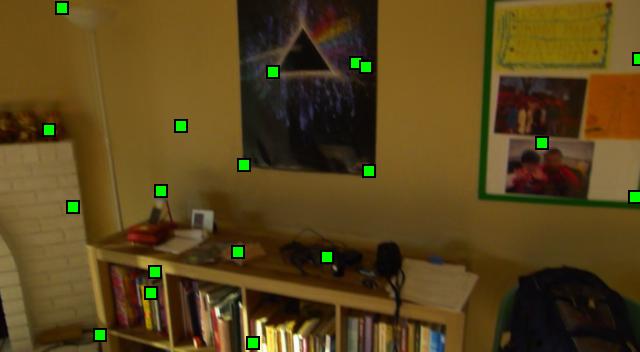}
\caption{[L=300] $\Delta R$~=~$1.51^\circ$, $\Delta t$~=~$6$cm}
\end{subfigure}
\hspace{1pt}
\begin{subfigure}{0.49\columnwidth}
\includegraphics[width=\textwidth]{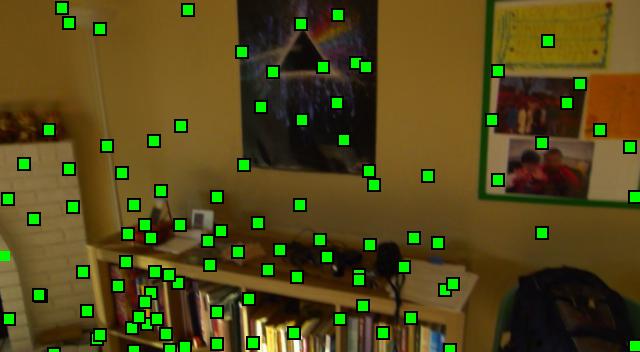}
\caption{[\blue{L=1000}] $\Delta R$~=~$0.64^\circ$, $\Delta t$~=~$2$cm}
\end{subfigure}
\\
(g) \hspace{8.4cm} (h)\\
\vspace{-2mm}
\caption{\textbf{Qualitative results on \textsc{Indoor-6}.} Detected scene landmarks are shown as green points on the images from scene1, and the rotation and translation errors in the SLD\textsuperscript{$\ast$} pose estimate are also reported below each image. (a)--(h) In all eight examples, the result on the left is for 300 landmarks, whereas the result on the right is for 1000 landmarks. Using 1000 landmarks instead of 300 landmarks produces more 2D--3D point constraints and the 2D locations are spatially better distributed in most images. which later yields a more accurate pose.}
\label{fig:qualitative}
\end{figure*}

The accuracy metric for SLD\textsuperscript{$\ast$} and SLD (both with 300 landmarks) is 50.8\% and 44.9\% respectively. This 6\% accuracy improvement of SLD\textsuperscript{$\ast$} can be attributed to better training labels generated using the proposed visibility estimation method. However, the best results of SLD\textsuperscript{$\ast$} is 70.1\% obtained using a 125$\times$8 ensemble trained on 1000 landmarks which is competitive with hloc~\cite{sarlin2019} at 71.4\%.

\customparagraph{Accuracy Speed Trade-off.}
Accuracy wise, SLD\textsuperscript{$\ast$} and hloc have similar performance on \textsc{Indoor-6}. Thus, we report a detailed accuracy and speed trade-off analysis for them. Figure~\ref{fig:scatterplot} shows that the two hloc configurations (where 15 and 20 matching pairs are used respectively) beats SLD\textsuperscript{$\ast$} by a small accuracy margin. However, these two hloc configurations are quite slow. The best SLD\textsuperscript{$\ast$} setting outperforms all other hloc configurations (which use 1, 2, 5 and 10 matching pairs). Moreover, even though 
smaller ensembles are slightly worse accuracy wise, they also run significantly faster. Note that, the reported timings are for sequential inference. In the parallel inference setting, SLD\textsuperscript{$\ast$} runs extremely fast because all the models are preloaded in GPU memory and all networks run inference in parallel. However, the memory footprint of the ensemble linearly increase with its size. Nonetheless, parallel inference may still be practical when sufficient GPU memory is available.

\customparagraph{Qualitative Results.} Finally, we present test images from scene1 in Figure~\ref{fig:qualitative} that were localized using two SLD\textsuperscript{$\ast$} models trained on 300 and 1000 landmarks respectively and also report the associated pose errors. These examples clearly demonstrate the benefit of scaling up the number of scene landmarks. The model for 1000 landmarks consistently produces more accurate results, due to more pose inliers being present and a better distribution of those inliers.
All SLD\textsuperscript{$\ast$} models were trained using NVIDIA V100 GPUs whereas queries were processed on a laptop with a RTX 2070 GPU. 
\section{Conclusion}

In this paper, we proposed SLD\textsuperscript{$\ast$}, an extension of the existing SLD framework for scene landmark detection-based camera localization. SLD\textsuperscript{$\ast$} is memory and storage efficient like SLD but it shows a dramatic improvement in performance (accuracy). The improvement makes SLD\textsuperscript{$\ast$} competitive with structure-based methods such as hloc~\cite{sarlin2019,sarlin2020superglue} while being about 40X faster. The improved accuracy can be attributed to two ideas proposed in the paper First, we proposed a new processing pipeline to generate more accurate training labels for training the detector. Secondly, we showed that partitioning the landmarks into smaller groups and training independent networks for each subgroup dramatically boosts accuracy when a large number of scene landmarks are present.
SLD\textsuperscript{$\ast$} is currently trained from scratch for each scene which is time consuming and expensive. Exploring ideas similar to those proposed recently for accelerating scene coordinate regression~\cite{BrachmannACE2023} could lead to faster training and is an important avenue for future work.

\nocite{*}
{
    \small
    \bibliographystyle{ieeenat_fullname}
    \bibliography{main}
}

\end{document}